\pdfoutput=1
\documentclass[11pt]{article}

\usepackage[]{EACL2023}

\usepackage{times}
\usepackage{latexsym}

\usepackage[T1]{fontenc}
\usepackage[utf8]{inputenc}

\usepackage{microtype}
\usepackage{graphicx}

\usepackage{inconsolata}
\newcommand{\comment}[1]{}
\newcommand{\nocomment}[1]{#1}
\newcommand{\COMMENT}[1]{}
\newcommand*{\mybox}[1]{\framebox{#1}}

\usepackage[russian,french,english]{babel}
\usepackage[utf8]{inputenc}
\usepackage[T2A,OT1]{fontenc}
\usepackage{tempora} % this supports Cyrillic

\newcommand\textcyr[1]{{\fontencoding{T2A}\selectfont #1}}

\newenvironment{enumeratorCompact}{\vspace{-1mm}
  \begin{enumerate}
    \setlength{\itemsep}{2pt}
    \setlength{\parskip}{0pt}
    \setlength{\parsep}{0pt}
  }
{ \end{enumerate}
  \vspace{-1mm}  }

\title{Linguistic Constructs as the Representation of the Domain Model \\in an Intelligent Language Tutoring System}

\author{Anisia Katinskaia\textsuperscript{1,2}, \ Jue Hou\textsuperscript{1,2}, \ Anh-Duc Vu\textsuperscript{1,2}, \ Roman Yangarber\textsuperscript{2} \\
        \textsuperscript{1} Department of Computer Science, University of Helsinki \\ 
        \textsuperscript{2} Department of Digital Humanities, University of Helsinki \\
        \texttt{first.last@helsinki.fi}}
\begin{document}
\maketitle
\begin{abstract}

This paper presents the development of an AI-based language learning platform Revita. It is a freely available intelligent online tutor, developed to support learners of multiple languages, from low-intermediate to advanced levels. It has been in pilot use by hundreds of students at several universities, whose feedback and needs are shaping the development. One of the main emerging features of Revita is the introduction of a system of {\em linguistic constructs} as the representation of domain knowledge. The system of constructs is developed in close collaboration with experts in language teaching. Constructs define the types of exercises, the content of the feedback, and enable the detailed modeling and evaluation of learning progress.

\end{abstract}
\section{Introduction}

The focus of this paper is a novel Domain Model of Revita,\footnote{revita.cs.helsinki.fi --- {\href{https://www.dropbox.com/s/jftlrr1ixcmg9az/eacl-demo.mp4?dl=0}{Link to a short demo here.}}} an Intelligent Tutoring System (ITS) for language learning~\cite{katinskaia-etal-2018-revita,katinskaia:2017-nodalida:revita}. The structure of Revita follows the classic design of ITS, with a Domain, Student, and Instruction model. The {\em Domain Model} describes what must be mastered by the learner: concepts, rules, etc.---known as {\em skills} in ITS literature---and {\em relationships} among them~\cite{wenger2014artificial,Polson1988foundations}. We represent the Domain Model as a system of {\em linguistic constructs}---a wide range of linguistic phenomena, including inflexion of various word paradigms, government relations, collocations, syntactic constructions, etc. The system of constructs is developed in collaboration with experts in language teaching. It impacts all components of Revita---the variety of exercises that it generates automatically, intelligent feedback, the modeling of learner knowledge and evaluation of learner progress.

The {\em Student model} represents the learner's proficiency. It is based on the history of answers given by the learner to exercises. 
The {\em Instruction model} embodies the pedagogical principles that lie behind the decisions: which exercises the learner is best prepared to do next, and which feedback should be provided to guide the learner toward the right answer.  The  models are interconnected in the ITS.

% In the recent years, distance learning, large-scale cross-border migration, the growth of international companies and communication---these and other trends increase the importance of and the need for flexible online language learning. Many commercial and academic projects focus on language learning, but most of them target beginners.  As for tools for students above the A2 CEFR level, we observe a scarcity of freely available resources beyond the prototype stage.  

Revita is currently piloted with real-world learners and teachers at several universities~\cite{stoyanova2021integration}. It is developed as a tool for learners and teachers of several languages: Finnish and Russian are currently the most developed languages; several ``beta'' languages including Italian, German, Swedish, Kazakh, Sakha, Tatar, English, Erzya, and others, are in early stages of development. The system is not meant to replace the teacher. For students, it provides 24/7 access to an unlimited amount of exercises for practice which fit the learner's current level, with immediate feedback and progress estimation. The platform also offers support for teachers: they can delegate the mundane work of creating hundreds of exercises as needed for each topic for students at different levels. It provides teachers with a range of instruments for sharing learning materials, working with groups of students, and for monitoring progress and evaluation.

The paper is organized as follows: Section~\ref{sec:prior-work} briefly reviews work on intelligent computer-assisted language learning (ICALL). The principles and ideas behind Revita are described in Section~\ref{sec:revita-principles}. It also describes its main components: linguistic constructs, automatic generation of exercises and feedback, and modeling of learner knowledge. Section~\ref{sec:tools} describes tools for learners and teachers.

\section{Prior Work}
\label{sec:prior-work}

Computer-aided language learning (CALL)\comment{emerged in the 1960s and} has been gaining interest with the rapid development of language technology. It is defined as ``the search for and study of applications of the computer in language teaching and learning''~\cite{levy1997computer}. %A broader definition of CALL is ``learning a language in any context with, through, and around computer technologies''~\cite{egbert2005call}. 
Applying ITS specifically to language learning and supporting CALL systems by intelligent and/or adaptive methodologies, such as expert systems (ES), natural language processing (NLP), automatic speech recognition (ASR)---defines the domain of intelligent CALL, or ICALL. The goal of ICALL is building advanced applications for language learning using NLP and language resources—corpora, lexicons, etc.~\cite{volodina2014flexible}.

The number of academic and commercial tools for language learning is growing drastically, e.g., the popular commercial systems like Duolingo, Rosetta Stone, Babbel, Busuu, iTalki, etc. Some CALL systems aim to give learners access to authentic materials~\cite{white2010theory}, the opportunity to interact with teachers and native speakers (e.g., the learning app {\em Lingoda} is a platform for live video classes), and provide text or sound feedback based on learner needs and knowledge~\cite{bodnar2017learner}. Modern CALL systems are also mobile, which increases their accessibility~\cite{derakhshan2011call,rosell2018autonomous,kacetl2019use}.

While some research points out that CALL systems, in their experiments, do not achieve increases in learner proficiency~\cite{golonka2014technologies,bodnar2017learner,rachels2018effects}, other work showed actual improvements in learner motivation and attitudes, retention of various learning concepts, communication between students and teachers, and overall language skills~\cite{yeh2019speaking,zhang2022types}. It has been suggested that in developing CALL system, pedagogical goals---rather than technological means---should be the primary focus~\cite{gray2008effective}.

\begin{figure}[t]
  \includegraphics[scale=0.21]{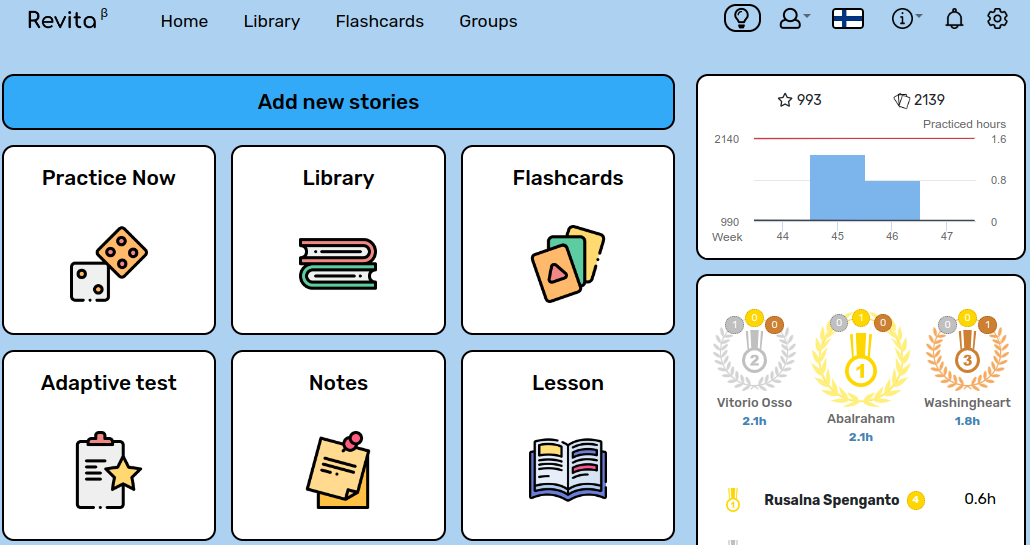}
  \caption{Revita's home page, with the main activities.}
  \label{fig:home}
\end{figure}

\section{Core Components of Revita}
\label{sec:revita-principles}

\subsection{Main Principles}

\begin{table*}[t]
\centering
\scalebox{0.90}{\begin{tabular}{ll}
\hline
\textbf{Constructs} & \textbf{Examples} \\
\hline
\textbf{Finnish} \\
(1) Necessive construction: Present                              & {\em Energiakriisin lähestyessä kaikki keinot \underline{on \textbf{otettava}} käyntiin.} \\ 
passive participle, with {\em -ttava} ending                 & (With the energy crisis approaching, all means \underline{must \textbf{be taken}} into action.) \\  
(2) Transitive vs. intransitive verbs                            & {\em Voisitko \underline{\textbf{sammuttaa}} valon?} (Could you \underline{\textbf{turn off}} the light?)\comment{{\em Voisitko \underline{\textbf{herättää}} minut huomenna?} (Could you \underline{\textbf{wake me up}} tomorrow?)} \\
(3) Verb government: translative case                            & {\em Kaupungit \underline{eivät ole muuttuneet\comment{muuttuvat} \textbf{energiatehokkaammiksi}}.} \\ 
                                                             & (Cities \underline{have not become \textbf{more energy efficient}}.) \\
(4) Present participle substitute for                            & {\em Maija \underline{kertoi \textbf{vanhempien asuvan}} kaupungissa.} \\
``that''-relative clause\comment{({\em Että-lauseenvastike}), with different subjects}                                     & (Maija \underline{said that \textbf{her parents live}} in the city.)\\[1ex]
\hline
\textbf{Russian} \\
(5) Verb: II conjugation                                         &  {\em \textcyr{Мы скоро \underline{\textbf{увидим}} восход.}} (We \underline{\textbf{will see}} the sunrise soon.) \\ 
(6) Pronoun: joint vs.~hyphenated\comment{Joint vs. hyphenated spelling}                      & {\em \textcyr{Нам нужно \underline{\textbf{кое о чем}} поговорить.}} (We need to talk \underline{\textbf{about something}}) \\
(7) Perfective vs. imperfective aspect                           & {\em \textcyr{Страны \underline{\textbf{согласовали}} проект о будущих отношениях.}} \\ 
                                                             & (The countries \underline{\textbf{agreed on}} a draft on future relations.) \\
(8) Dative subject \& impersonal verb                             & {\em \textcyr{\underline{\textbf{Мне} необходимо поговорить} с врачом.}} (\underline{\textbf{I} need to talk} to a doctor.) \\ [1ex]
\hline
\textbf{German} \\
(9) Past perfect tense                                           & {\em Ich \underline{\textbf{wäre}} mit ihm \underline{\textbf{gekommen}}, aber er wurde krank.}\\
                                                             & (I \underline{\textbf{would have come}} with him, but he got sick.) \\
(10) Weak masculine nouns                                         & {\em Ich möchte \underline{\textbf{den Jungen}} kennenlernen.}  (I want to meet \underline{\textbf{the boy}}.) \\
(11) Prepositions governing dative case                       & {\em Wir sind \underline{\textbf{aus dem} Haus} gelaufen.} (We ran \underline{\textbf{out of the} house}.) \\

\end{tabular}
}
\caption{Examples of \underline{\em grammatical constructs} found in sentences (underlined). 
{\em \textbf{Candidates}} are words that will be chosen for exercises about the constructs (marked in bold).}
\label{constructs}
\end{table*}

Revita's approach is founded on the following primary principles:
\begin{enumeratorCompact}
    \item {\em Practice should be based on authentic content}: the learner (or teacher) can upload any text from the Internet or a file to the system.
    \item {\em Exercises are automatically generated}, based on any authentic text chosen by the learner.
    \item {\em Exercises are personalized}: they match the learner’s current skill levels, so that each new exercise is selected to be a challenge that the learner is ready to meet.
    \item {\em Immediate feedback}: rather than saying only ``right/wrong'', the tutor {\em gradually guides} the learner toward finding the correct answer by providing hints, which begin as general hints and give more and more specific information about the answer.
    \item {\em Continual assessment} of skills allows the tutor to select exercises optimally personalized for each learner based on past performance.
\end{enumeratorCompact}

The first principle is the bedrock of the philosophy behind Revita---the story-based approach. All learning activities are based on authentic texts, which should be inherently interesting for the learner to read and motivate her to practice longer. A few sample texts are available in the system's ``public'' library for each language; also, several new stories are recommended as so-called ``Stories of the day''---which are crawled daily from several websites. But the main idea is that texts be selected and uploaded by the learners themselves (or teachers); see Figure~\ref{fig:home}, the button ``Add new stories'' allows one to upload new material into Revita.

\begin{figure*}[t]
  \hspace{-0.3cm}
  %%                    left bot right top
  \includegraphics[trim=0mm 37mm 0mm 0mm, clip, scale=0.4]{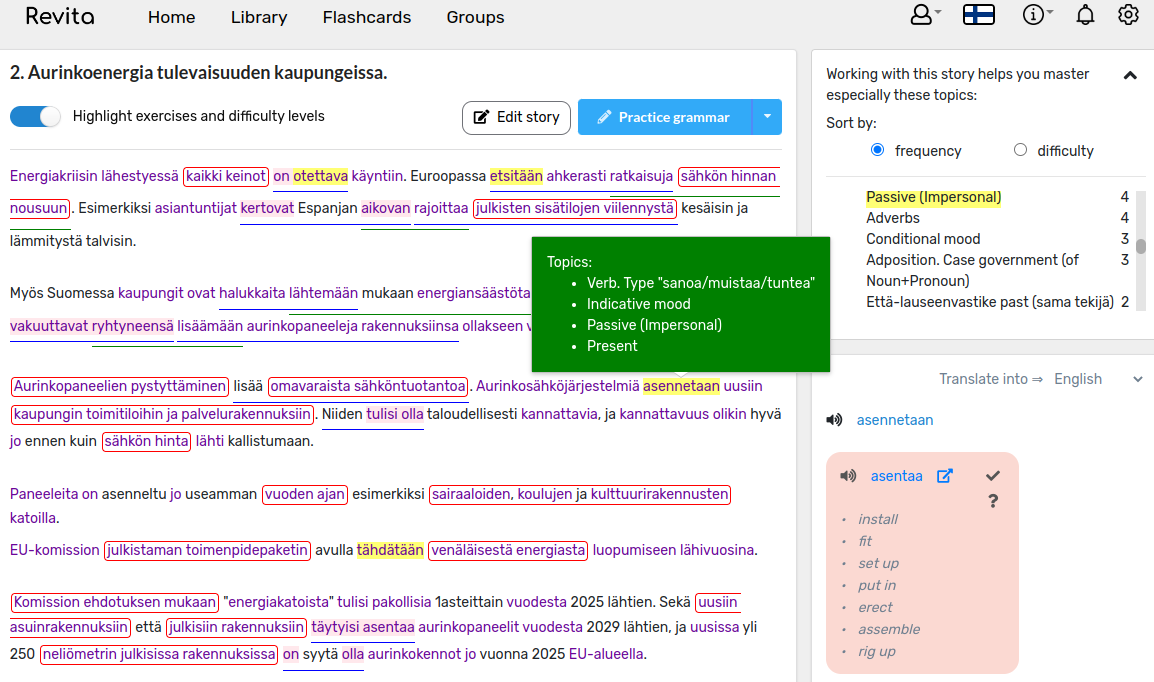}
  \caption{Preview mode for a story (before practice). 
  All violet words may appear in an exercise.
  Noun phrases and prepositional phrases are circled in red. 
  All government relations and constructions are underlined.
  Top-right corner---the list of constructs found in the story.
  Bottom-right corner---translation of the clicked word: ``asennetaan'', into English (target language can be selected). 
  The green box over the clicked word lists all constructs related to it.}
  \label{fig:read}
\end{figure*}

\subsection{Linguistic Constructs}

The central aspect of Revita's approach is the system of linguistic constructs that are represented in the Domain model. {\em Constructs} are linguistic phenomena or rules, that vary in specificity: e.g., a construct (in Finnish) may be {\em verb government}: the verb {\em tutustua} (``to become acquainted'') requires its argument to be in the illative case ({\em ``into something''}), while {\em tykätä} (``to like'') requires its argument in the elative case ({\em ``from something''\comment{jostakin}}), etc. Constructs also include all {\em constructions}, as construed in Construction Grammar (CG). CG treats many phenomena---grammatical constructs, multi-word expressions (MWEs), collocations, idioms, etc.---within a unified formalism. Examples of constructs for several languages are shown in Table~\ref{constructs}. 

When customizing the system for a new language, we engage experts in language teaching in creating the inventory of constructs, which need to be learned by students. %, and connecting them together into a set of lessons. {we don't mention lessons in this paper, expect future work}
Currently, Finnish and Russian have the most developed system of constructs, each with over 200 constructs. (Potentially, the number can be much larger.)
The Russian construct system evolved from the grammatical constructs covered in tests for L2 learners developed at the University of Helsinki over 20 years. 
The Finnish constructs are based on inventories of grammatical topics developed by experts in teaching Finnish as L2.  

As shown in the examples in Table~\ref{constructs}, each construct needs to be identified in the text. For this we use HFST morphological analyzers, neural dependency parsers, and rule-based pattern detection.

In Example (1), for construct ``Present passive participle with {\em -ttava} ending,'' the rule matches the participle {\em ``otettava''} by morphological features: participle, present tense, passive voice. This is then recognized as the head of the ``necessive'' construction {\em ``on otettava''} (``must be taken''), detected by a rule that matches: the singular 3rd person present form of modal verb {\em ``olla''} (``to be''), and the present passive participle, in the nominative case.

In Example (2), the construct ``Transitive vs.~intransitive verbs'' is detected by using dictionaries of verb lemmas or by rules that detect regular ending patterns in verb lemmas 
{(e.g., {\em sammu\textbf{ttaa}} vs {\em samm\textbf{ua}}, ``turn something off'' vs. ``turn itself off'')}.

\begin{figure*}[t]
  \includegraphics[scale=0.4]{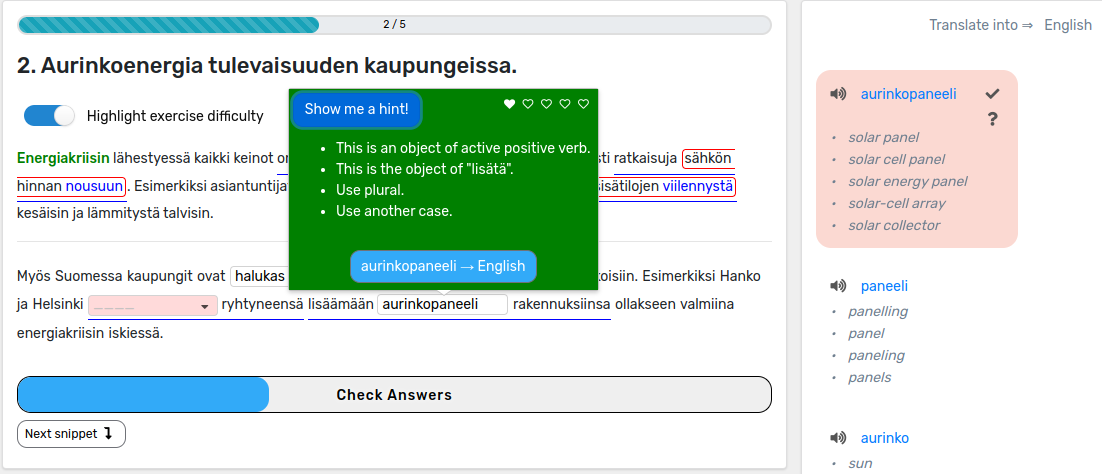}
  \caption{Practice mode with a story. Figure shows the second paragraph of a story with three exercises: two clozes (``halukas'' and ``aurinkopaneeli'') and one MC. Previous answers are marked green and blue---correct and incorrect. The green box shows the set of opened hints for a cloze exercise.}
  \label{fig:practice}
\end{figure*}

Verb government relations are detected by several components: large sets government patterns (2000-3000 per language); pattern matching of noun phrases, prepositional phrases, and analytic verb forms; dependency relations detected by parsers. 
In Example (3), a government pattern for the \nocomment{intransitive} verb {\em ``muuttua''} (``to change'') requires an argument in translative case---here, the comparative adjective {\em energiatehokkaammiksi} (``more energy-efficient''). 
The government detector will find an argument of {\em ``muuttua''} regardless of its position in the sentence, and for any form of the verb, including complex analytic forms, e.g., the negative perfect tense {\em ``eivät ole muuttuneet.''}

Detecting longer and more complex syntactic constructions relies on all of the components mentioned above. 
In (4), to match the construction {\em ``kertoi vanhempien asuvan''}, the government rule states that the verb ``kertoi'' (``said'') must govern a subordinate clause starting with {\em ``että''} (``that''); the {\em substitute} clause contains a noun phrase in the genitive case, which acts as the subject ({\em ``vanhempien''}) and a genitive active participle ({\em ``asuvan''}).

The user can preview all constructs identified in a story in the Preview Mode prior to practice, see Figure~\ref{fig:read}. All noun and preposition phrases are circled, government relations and syntactic constructions are underlined.  A list of all constructs found in the story is in the top-right: one can click on the list to highlight all instances of the construct found in the story. Clicking on any word in the story will also show all constructs linked to it in a green box, above the clicked word. This lets the learner (or teacher) see what can be exercised in the given text.

\subsection{Exercise Generation Based on Constructs}

Revita offers several practice modes; the main activity is the Grammar Practice Mode based on a story, see Figure~\ref{fig:practice}. 
Revita offers ``cloze'' (fill-in-the-blank) and multiple-choice (MC) exercises. A cloze exercise is shown as a text box, with the lemma of the expected answer given as a hint to the learner. In Figure, the lemma in the box is \mybox{{\em aurinkopaneeli}} (``sun panel''). The learner is expected to insert the correct form of this word that suits the context; here, it is the plural partitive case ({\em ``aurinkopaneeleja''}). Each word picked to be exercised must be disambiguated---we have to know the correct lemma to show to the learner. Disambiguation is performed by agreement rules and by dependency parsers. For analytic verb forms, such as {\em ``on otettava''} (``should be taken''), the cloze box will show the lemma of the {\em head} verb: \mybox{{\em ottaa}} (``to take'').

All {\em candidates}---potential exercises in practice mode---are based on the detected constructs. In Example (3) for Finnish in Table~\ref{constructs}, an exercise on  the construct ``Verb government'' is in bold: the learner will see the lemma \mybox{{\em energiatehokas}} (``energy-efficient''). To insert the correct form in the translative case, the learner needs to know which case is required by the governing verb.  MC exercises are more targeted: the options---knowns as ``distractors''---to choose from are generated based on the exercised construct.
Therefore, the same word or construction may have more than one set of distractors, since more than one construct may be linked to the candidate. 

Distractors are created by rules and morphological generators. In Example (6), for the construct ``Pronoun: joint vs.~hyphenated spelling" a rule generates distractors like: {\em \textcyr{``кое о чем''}}, {\em \textcyr{``кое-о-чем''}}, {\em \textcyr{``о кое-чем''}} (``about something''). For transitive vs.~intransitive verbs, Revita uses dictionaries of lemma pairs\comment{or generates distractors' lemmas using a rule}. However, the distractors must be {\em inflected} forms that fit the context, not lemmas. We use morphological generators to produce the required inflected forms\comment{ based on given lemmas and sets of grammatical tags}.

MC distractors are often an effective way of learning a particular construct. In Example (4), e.g., the construction requires the subject to be in genitive case; it is useful to offer the lemma {\em ``vanhemmat''} (``parents'') in other cases (nominative, partitive, etc.) These forms, which differ only by case, are produced by the morphological generator.

\subsection{Feedback}

Feedback is a second essential feature in Revita. The learner gets {\em multiple attempts} for every exercise. Feedback is designed to gradually guide the learner toward the correct answer by providing a sequence of hints that depend on the context, the constructs linked to the exercise, and on the answer that was given by the learner.
Hints are ordered so they become more specific on subsequent attempts.  For example, if a verb governs the partitive case, the feedback sequence may be: {\em ``The is the object of the verb 'lisätä'.''} \textrightarrow {\em ``Use another case.''} \textrightarrow {\em ``Use partitive case.''}
The learner can also request hints {\em before} inserting an answer: as shown in the green box in Figure~\ref{fig:practice}, four of the available hints are already ``used up,'' (one white heart remaining). Requesting hints indicates that the learner has not mastered the concept, and affects the learner's scores.
Feedback that depends on the context gives information on whether a word in question is part of some construction or relies on a governing head (verb, noun, or adjective), etc. 
Hints also appear as {\em underlining} of syntactically related elements in the context.\comment{: this information is available after analyzing a story, as well as other grammatical features (case, number, tense, etc). }

When the learner inserts an answer which does not match the expected one (the same as in the original story), Revita analyses the answer and checks which grammatical features are not correct. To give feedback on these features in the order of increased specificity, Revita uses a language-specific hierarchy of features. 
For example, in Russian, the hierarchy specifies that the hint about an incorrect gender of an adjective are shown before hints about incorrect number or case.\comment{---noun lemmas are given and their gender cannot be modified. At the same time, a gender hint is shown  for adjectives.}

Some feedback messages are generated in the stage when the construct is mapped to a text. For example, for an exercise with the participle {\em ``asuvan''} in {\em substitute that-clause} construction (see example (4) in Table~\ref{constructs}), we generate the feedback: {\em This is equivalent to ``...kertoi että vanhemmat asuvat...''} (``...said that parents live...'')---by generating the actual clause which is substituted by the participle. To generate this feedback, Revita uses information about the syntactic roles of each word in the original construction {\em ``kentoi vanhempien asuvan''}, and the required grammatical features of the forms in the feedback---to get these forms, we use the morphological generator. 

All mechanisms which define the order and the content of feedback hints and algorithms of sampling exercises for students are part of the Instruction Model of Revita.

\subsection{Learner Modeling and Exercise Sampling}

All learner answers and all requested hints to each exercise are recorded. A learner may attempt to answer each exercise multiple times. For each attempt, Revita analyzes the answers and the requested hints to calculate {\em credits and penalties} for the corresponding language constructs. 

The collected information on performance with constructs is used to model the learners' skill and the difficulty of the constructs.
To model students' skills and exercise difficulty, we employ Item Response Theory (IRT)~\cite{embretson2013item,van2013handbook}. IRT comes from psychometrics and has a wide range of applications in education~\cite{klinkenberg2011computer}.
The {\em Item} in IRT is a task that the learner should solve. Most IRT applications have a clear definition of an {\em item}, and a clear credit standard. 
The classic example of an item is an test question in mathematics: it is unambiguous and there is a clear judgment of the answer---correct or wrong.
Our major challenge is that language constructs are not directly judged, as test items in other learning domains.\comment{ (such as mathematics, physics, etc.)} 
It is challenging to determine the credit and penalty for each construct based on the student's answer, % to an exercise,
because the link from exercise to concepts is {\em one-to-many}.\comment{ and requested hints.}

To date, we have collected 570K answers for Russian exercises. Experiments with this data show a strong correlation between students' proficiency, as estimated by IRT and by the teachers. \nocomment{This suggests that IRT is able to converge on a reliable estimate of learner proficiency.} The difficulty estimation of exercises, Interestingly, the estimates of exercise difficulty do not correlate with teachers' judgments, which agrees with the findings of other researchers~\cite{lebedeva-2016-placement-test}.  

Exercises are sampled for practice based on the difficulty of the hardest construct linked to each exercise.\footnote{At present, we assume that the difficulty of an exercise depends on the {\em hardest} construct linked to it.} The difficulty of constructs is modeled by IRT. We aim to provide exercises that are best suited to each student's proficiency levels. For each possible exercise, IRT first estimates the probability that the student will answer the exercise correctly---then the probability of picking this exercise for practice is sampled from a normal distribution around the mean of a $50\%$ chance that the learner would answer correctly.  Thus, on average, the exercises are not too difficult and not too easy.

For languages with insufficient learner data for training IRT, we ask teachers to assign manually CEFR difficulty levels to constructs.  Earlier experiments using specialized ELO ratings for assessing learner skills and evaluating the difficulty of linguistic constructs are presented in~\citet{hou-etal-2019-modeling}. 

\section{Tools for Students and Teachers}
\label{sec:tools}

At any time, the student can set her CEFR proficiency level manually, or to take an {\em adaptive} placement test to estimate proficiency. 
The test draws on a bank of questions prepared by teachers; the sampling of questions is driven by an IRT model trained on learner data.
\comment{by teachers and each question is marked by a CERF level.}
After that, the estimate of the learner's prodiciency levels are adjusted according to the correctness of answers to exercises.

The learner can upload a story from any website or a local file. 
To each uploaded text, Revita applies classification by semantic topic---culture, science, sport, politics---and difficulty classifiers.
\comment{that run over each text that is added to the platform.

After that, the student is offered a variety of learning activities. 

there is also an option to practice with a random story with a defined topic}

The {\em Preview mode} (see Figure~\ref{fig:read}) allows the user to read a story, edit it in case of inaccuracies, and review the grammatical topics that can be learnt through practicing with this story. Clicking on each word provides its translation into a number of languages. 
\comment{}{The learner can mark if she knows a word or not.}
All unknown words are added to the learner's personal set of flashcards, which are used for Vocabulary Practice. 

The {\em Practice mode} presents the grammar and listening exercises---the learner can hear a segment of text in context, and is expected to type the answer correctly in the practice box. The user can also practice with a story in the Competition Mode, against a bot: the difference from the standard practice is that the learner needs to do the exercises faster than a bot---whose skill levels approximately matches those the learner. Another option is to practice with a Crossword built on the authentic text---the translations of words are used as hints. 

Revita offers various statistics and info-graphics to track progress on grammar constructs and vocabulary. These analytics are available to the learner and to the teacher. Revita allows teachers to build groups of students, share texts with them, and create tailored exercises that can be shared with the group. Revita allows teachers to track how their students practice and how well they perform on  various tasks.

\section{Conclusions and Future Work}

This paper presents an in-depth discussion of the novel core component of the Revita language learning system---the Domain model embodied in a system of linguistic constructs. This system of constructs shapes all aspects of the learning experience in Revita and improves the quality of exercises and feedback. It also enables the modeling of learner knowledge more accurately, to provide informative progress analytics, and to offer exercises most appropriate for the learner's current level.

We have results from pilot studies with Finnish and Russian L2 learners using the new Domain Model, but the discussion of the results is beyond the scope of this paper. In the future, we will improve the Domain Model by adding more information about the {\em interactions} and dependencies among the constructs---which will enable the creation of more intelligent learning paths. We will also add new types of activities, e.g., speech exercises. 
\comment{We are also working on developing a new mode of interaction with Revita, i.e., a system of lessons---sets of grammatical and lexical topics that can be practiced together in a form of different activities.
}

% not counted within 6 pages
\section*{Limitations}

Revita works with many languages, however, at present, only Finnish and Russian have a sufficiently developed inventory of constructs that they can be actually used by students in real-world scenarios.
Most other languages have a limited set of constructs, which affects the quality and variety of the exercises, as well as limited feedback. 
Developing a substantial inventory of constructs is a complex task, that rerquires expertise in computational linguistics, as well as in language pedagogy.
As mentioned above, Finnish and Russian have on the order of 200 constructs. Meanwhile, ``the Great Finnish Grammar'' has over 1500 articles \cite{vilkuna-2004-iso}, each of which introduces at least one construct, which, in principle, constitutes an aspect of the linguistic competency of a native speaker. A fascinating research challenge is determining the ``essential'' core inventory of constructs, which can support effective learning. Our experience so far with the rather modest inventories suggest that they already bring enormous value to learners and teachers~\cite{stoyanova2021integration}.

The approach relies on arbitrary authentic texts being uploaded from the web; sometimes these texts cannot be extracted from the web site without some inaccuracies. Also, the original texts may contain typos, mistakes, etc. These problems should be fixed manually by editing the text. Of course, learners with a low proficiency level cannot do that independently. To avoid having these mistakes  negatively affect the learning, the stories can be checked by a human teacher / tutor. 
We also plan to employ strong language models for grammatical error detection to identify such potential problems and highlight them to alert the user that additional checking may be needed.   

Revita relies on the text when checking the learner's answers. Currently, only one correct answer allowed---the one that is present in the story. Sometimes the wordform entered by the learner may also be valid in the given story context---``alternative correct'' answers. In such cases, Revita may still tell the learner that the answer is not correct. 
This is one of the important problems that we are researching at present, using neural models for detection of grammatical errors \cite{katinskaia-2019-ACL-multi-admiss,katinskaia2021assessing}.

Revita also has certain limitations related to the use of external tools and services: dependency parsers, morphological analyzers, and external dictionaries---all may contain inaccuracies and errors. All of these factors can be a source of mistakes in the intelligent tutor: wrong analyses, incorrectly disambiguated lemmas, missing translations, etc.
The system tries to collect {\em multiple sources of evidence} for its predictions to raise the confidence in---and precision of---the predictions.
When the confidence is low---e.g., in the presence of conflicting evidence---the exercise, feedback, etc., is {\em not} offered to the learner.

% EACL 2023 requires all submissions to have a section titled ``Limitations'', for discussing the limitations of the paper as a complement to the discussion of strengths in the main text. This section should occur after the conclusion but before the references. It will not count towards the page limit.

% The discussion of limitations is mandatory. Papers without a limitation section will be desk-rejected without review.
% ARR-reviewed papers that did not include ``Limitations'' section in their prior submission, should submit a PDF with such a section together with their EACL 2023 submission.

% While we are open to different types of limitations, just mentioning that a set of results have been shown for English only probably does not reflect what we expect. 
% Mentioning that the method works mostly for languages with limited morphology, like English, is a much better alternative.
% In addition, limitations such as low scalability to long text, the requirement of large GPU resources, or other things that inspire crucial further investigation are welcome.

% not counted within 6 pages
\section*{Ethics Statement}

Revita is designed to carefully guard the privacy of its users---learners and teachers. It does not share any personal information collected during the learner's practice with any third parties. The teacher can track the learner's performance only if the learner has explicitly accepted the invitation to join the teacher's group.

Any authentic text material uploaded into the system is visible only in the user's personal {\em private} library.  If the teacher shares a story with a group of students, it is visible only inside the group library, never to anyone outside the group. Texts pre-loaded into Revita's public library come either from sources that have given us explicit permission to use their content, or from the public domain.
\comment{only separate short paragraphs from texts.}

% Scientific work published at EACL 2023 must comply with the \href{https://www.aclweb.org/portal/content/acl-code-ethics}{ACL Ethics Policy}. We encourage all authors to include an explicit ethics statement on the broader impact of the work or other ethical considerations after the conclusion but before the references. The ethics statement will not count toward the page limit (8 pages for long, 4 pages for short papers).

\section*{Acknowledgements}

\bibliography{revita}
\bibliographystyle{acl_natbib}

\end{document}